\documentclass[letterpaper, 10 pt, conference]{ieeeconf}  

\IEEEoverridecommandlockouts                              

\usepackage{cite}
\usepackage{subfigure}
\usepackage{amsmath,amsfonts,amssymb}
\usepackage{xcolor}
\usepackage{optidef}
\usepackage{bm}
\usepackage{algorithm}
\usepackage{algpseudocode}
\usepackage{multirow}
\usepackage{multicol}
\usepackage{todonotes}
\usepackage{hyperref}

\renewcommand{\vec}[1]{\bm{#1}}
\newcommand{\mat}[1]{\mathrm{\bm{#1}}}

\title{\LARGE \bf
Simultaneous Contact Location and Object Pose Estimation Using Proprioception and Tactile Feedback
}

\author{Andrea Sipos$^{1}$ and Nima Fazeli$^{1}$
\thanks{$^{1}$The authors are with the Department of Robotics at the University of Michigan, 2505 Hayward Drive, Ann Arbor, USA
        {\tt\small <asipos,nfz>@umich.edu}}%
}

\begin{document}

\maketitle


\begin{abstract}
Joint estimation of grasped object pose and extrinsic contacts is central to robust and dexterous manipulation. In this paper, we propose a novel state-estimation algorithm that jointly estimates contact location and object pose in 3D using exclusively proprioception and tactile feedback. Our approach leverages two complementary particle filters: one to estimate contact location (CPFGrasp) and another to estimate object poses (SCOPE). We implement and evaluate our approach on real-world single-arm and dual-arm robotic systems. We demonstrate that by bringing two objects into contact, the robots can infer contact location and object poses simultaneously. Our proposed method can be applied to a number of downstream tasks that require accurate pose estimates, such as tool use and assembly. Code and data can be found at \url{https://github.com/MMintLab/scope}.
\end{abstract}

\section{INTRODUCTION}
\label{sec:intro}

Accurate localization of grasped objects and their contacts with the environment is an important step towards robust and dexterous object manipulation. Inaccurate object pose estimation leads to a number of potential downstream failures including grasp instability, finger misplacement, and unreliable interactions with the environment. These failures significantly impact downstream task performance for applications ranging from robotic tool use to part assembly.

Current state-of-the-art techniques primarily use visual feedback with analytical or data-driven models to identify and localize objects in a scene \cite{kokic2019learning, tremblay2018deep, tremblay2020indirect, wen2020robust, romero2013non, du2021vision, honda1998real, haidacher2003estimating, li2020hand, paul2021object}. The limitation of these approaches is the inevitable occlusion of objects during grasping and manipulation, which significantly deteriorates estimation performance. Further, most existing techniques are one-shot (i.e., the scene is perceived once and an inference is made.) Beyond these limitations, there are many instances of manipulation where visual feedback is not well-suited to the task; for example, rummaging in a bag or using a tool in a tight space \cite{zhong2021tampc, zhong2022soft}. Recent progress in collocated tactile sensing \cite{bauza2020tactile, ma2021extrinsic, yamaguchi2019tactile, bimbo2016hand, zwiener2018contact, fan2022enabling, alvarez2017tactile} enables state-estimation using tactile information, but only at the cost of significant hardware overhead. Many novel tactile sensors employ deformable components that fatigue, necessitating replacement parts after regular use. Further, several of these tactile sensors are vision-based which increases data-processing complexity to include streams of high-resolution images. Our method could be used with these sensors, but does not require them.

\begin{figure}
    \centering
    \includegraphics[width=0.48\textwidth]{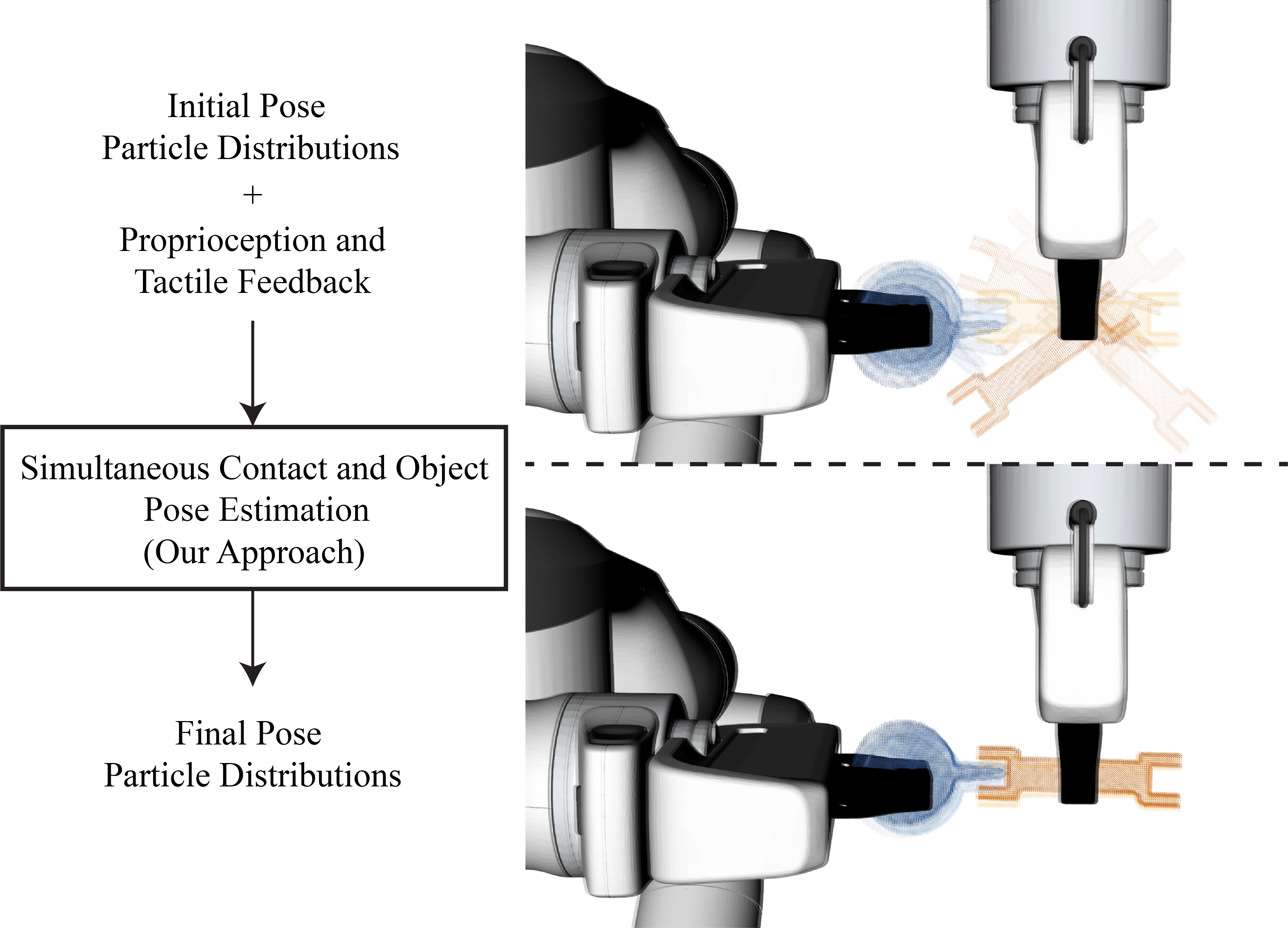}
    \caption{Two robots grasping two tools with high pose uncertainty. The top pane shows samples drawn from the initial distribution of poses. Our method uses contact between the two objects to iteratively update object poses and contact location. Bottom pane shows samples from the posterior distribution over object poses
    where the robots have successfully localized the objects and their contact location.}
    \label{fig:opp_teaser}
\end{figure}

Instead, our method leverages proprioceptive feedback from robot joint angles and torques and tactile feedback from 6DOF force-torque sensors at the wrist to update two complementary particle filters over contact location and object pose. We implement our method on real-world single and dual-arm robot systems to demonstrate the efficacy of our approach in estimating contact location and object pose. An overview of our method can be seen in Fig.~\ref{fig:opp_teaser}

\section{RELATED WORK}
\label{sec:related_work}

In-hand object pose estimation has a long history in robotics. The literature can be broadly categorized into visual and tactile/distributed in-hand object pose estimation. There are also overlaps of visuotactile state-estimators that we discuss below.

\textbf{Vision-based:} A large portion of the literature has focused on vision-based techniques that attempt to estimate grasped object poses despite significant occlusions \cite{kokic2019learning, tremblay2018deep, tremblay2020indirect, wen2020robust, romero2013non, du2021vision, honda1998real, haidacher2003estimating, li2020hand, paul2021object}. These approaches use RGB-D sensing and have the advantage of leveraging recent advances in learning to mitigate the need for object geometric models, instead relying on learning to generalize to novel scenes. Two important challenges to address for these methods are: i) reasoning over external contacts, especially when the interaction is significantly occluded and ii) multi-object interactions and pose resolution. Our approach is complementary to these methods for resolving precise poses using extrinsic contacts.

\textbf{Tactile and Visuotactile:} These approaches use collocated tactile sensing at the interface between the robot and object \cite{bauza2020tactile, ma2021extrinsic, yamaguchi2019tactile, bimbo2016hand, zwiener2018contact, fan2022enabling, alvarez2017tactile, manifoldpf, hapticrender, probingpf, rbpf, contactformations}. This collocated sensing can be provided by localized force-torque sensors, vision-based tactile sensors, or prototypes of robot ``skin''. A number of methods have also been developed that combine vision and touch \cite{bimbo2012object, von2021precise, chavez2018contact, ding2018hand, honda1998real, haidacher2003estimating, dikhale2022visuotactile, chaudhury2022using, multisensor} to exploit the complementary nature of these modalities. Our approach can complement these methods by additionally integrating proprioception from the robot (or articulated fingers when available) to further refine the pose estimates. More specifically, our proposed approach builds on \cite{manuelli2016localizing}, extending the contact particle filter to the grasped object and simultaneously resolving object pose and external contact. Further, we show how our methodology can be extended to dual-arm systems where the pose of two objects and their contact locations can be resolved simultaneously. Our method is a type of approach that can augment the collocated tactile sensors and vision-based techniques discussed above.

\section{METHODOLOGY}
Our method maintains and updates a joint probability distribution over the estimated contact location on each object as well as its in-hand pose by building on the Contact Particle Filter (CPF) proposed by \cite{manuelli2016localizing}. In the following, we first present our assumptions. Next, we briefly describe the CPF. Then, we present the details of our generalization to CPF for Grasped Objects (CPFGrasp). Finally, we present the Simultaneous Contact and Object Pose Estimation (SCOPE) method.  

\subsection{Assumptions}

Here, we assume that objects are rigid and that geometric models are provided. Further, we assume that the robot can estimate externally applied wrenches using joint torques or external force and torque sensing. We assume that all contacts are point contacts and that no moments are applied at the contact location.

\subsection{Contact Particle Filter Review}

The CPF was introduced by \cite{manuelli2016localizing} as a tractable method of locating externally applied contact forces on a rigid-body robot. The premise of this method is that the joint torques due to external forces acting on the robot $\vec{\tau}_{ext}$ can be explained by a contact wrench $\vec{\Gamma}_c \in \mathbb{R}^6$ and its corresponding location of application on the robot's surface $\vec{r}_c \in \mathbb{R}^3$. The authors assume that all contacts are point contacts and subsequently that the torque component of $\vec{\Gamma}_c$ is zero. Both the force component of $\vec{\Gamma}_c$ and the contact location $\vec{r}_c$ must be estimated. Here, we briefly introduce the filter and refer the reader to \cite{manuelli2016localizing,manuelli2018localizing} for further details. 

Let $\vec{q} \in \mathbb{R}^n$ denote the configuration of the rigid-body robotic system with $n$ joints. Then, the dynamics of the robot are described by:
\begin{equation}
\label{eomrigidbody}
    \mat{H}(\vec{q})\ddot{\vec{q}} + \mat{C}(\vec{q},\dot{\vec{q}})\dot{\vec{q}} + \vec{g}(\vec{q}) = \mat{B}\vec{\tau} + \vec{\tau}_{ext}
\end{equation}

\noindent where $\mat{H}(\vec{q}) \in \mathbb{R}^{n \times n}$ is the inertia matrix, $\mat{C}(\vec{q},\dot{\vec{q}}) \in \mathbb{R}^{n \times n}$ is the Coriolis and centrifugal terms, $\vec{g}(\vec{q})$ is the conservative components, $\mat{B}$ maps the motor torques to joint torques, $\vec{\tau}$ is the motor torques, and $\vec{\tau}_{ext}$ is the external torque acting on the system. 

The residual joint torques $\vec{\gamma}$ experienced by the robot due to external contacts can be computed using the residual observer proposed by \cite{de2005sensorless}:

\begin{equation}
    \label{eq:residualobserver}
    \gamma(t) = 
    \mat{K}_{I}\Big(\mat{H}(\vec{q})\dot{\vec{q}} - 
    \int_{0}^{t} \big(\mat{B}\vec{\tau} + 
    \mat{C}^T(\vec{q}, \vec{\dot{q}})\vec{\dot{q}} + 
    \gamma(s) \,ds\big)\Big)
\end{equation}

\noindent where $\mat{H}(\vec{q})\dot{\vec{q}}$ is the generalized momentum of the robot and $\mat{K}_{I} > 0$ is a diagonal gain matrix.

The contact wrench $\vec{\Gamma}_c$ and the point of application on the surface of the robot $\vec{r}_c$ can be inferred from the residual joint torques $\vec{\gamma}$ by solving the following optimization:

\begin{mini}|s|
{\vec{r}_c, \; \vec{\Gamma}_c}{ (\vec{\gamma}-\mat{J}^T_c(\vec{q})\vec{\Gamma}_c)^T (\vec{\gamma}-\mat{J}^T_c(\vec{q})\vec{\Gamma}_c)}
{}{}
\addConstraint{\vec{r}_c \in \mathcal{S}_R, \; \vec{\Gamma}_c \in \mathcal{F}_c}
\label{eq:cpfbaseoptim}
\end{mini}

\noindent where $\mat{J}_c(\vec{q})$ denotes the geometric Jacobian of point $\vec{r}_c$, $\mathcal{S}_R$ denotes the set of points belonging to the surface of the robot, and $\mathcal{F}_c$ denotes the friction cone. 

This optimization is non-convex; however, \cite{manuelli2016localizing} observes that given a hypothesis for $\vec{r}_c$ and the polyhedral friction cone approximation from \cite{stewart2000implicit}, the resulting program is quadratic. Using this insight, \cite{manuelli2016localizing} proposes the CPF as a method to iteratively solve for the contact location and wrench. Specifically, for a choice of $\vec{r}_c$, they solve the following optimization:

\begin{mini}|s|
{\vec{\alpha}}{ (\vec{\gamma}-\mat{J}^T_c(\vec{q})\vec{\Gamma}_c)^T {\Sigma_{m}^{-1}} (\vec{\gamma}-\mat{J}^T_c(\vec{q})\vec{\Gamma}_c)}
{}{\text{QP}(\vec{\gamma}|\vec{r}_c)=}
\addConstraint{\vec{\alpha} \geq 0, \quad \vec{\Gamma}_c = \sum_{i=1}^{N_{f}}} \alpha_{i} \vec{f}_{c, i}
\label{eq:cpfquadoptim}
\end{mini}

\noindent where $N_{f}$ is the order of the polyhedral friction cone approximation, $\Sigma_{m}$ is calibrated sensor noise, and the constraint on $\vec{\alpha}$ indicates that contacts can only push against the surface. The probability that candidate contact location $\vec{r}_c$ explains the residual joint torques $\vec{\gamma}$ is then written as:

\begin{equation}
    \label{eq:cpf_score}
    p(\vec{\gamma}| \vec{r}_c) \sim \exp(-\frac{1}{2}\text{QP}(\vec{\gamma}|\vec{r}_c))
\end{equation}

By sampling a set of surface points and using a random noise model in lieu of a motion model, the particle filter iteratively improves the estimate of the contact location and the wrench applied at that location. This approach is summarized in Alg.~\ref{alg:cpf}, where $\mat{R}_c$ is the belief distribution over the contact location represented as a particle set.

\begin{algorithm}
\caption{CPF}\label{alg:cpf}
\begin{algorithmic}
\Procedure{CPF}{$\vec{\gamma}$}
\State $\mat{R}_c \gets \{\vec{r}_{c,0}, \cdots, \vec{r}_{c,N_{clp}} \} \in \mathcal{S}_R$ \Comment{Init particles}
\State $N_{cs} \gets $ Set Number of Iterations
\For{$i \gets 0; \; i < N_{cs}; \; i++$}
    \State $\mat{R}_c \gets $ Noise-Model$(\mat{R}_c)$
    \For{$\vec{r}_c$ in $\mat{R}_c$}
        \State $p(\vec{\gamma} | \vec{r}_c) \gets \text{QP}(\vec{\gamma} | \vec{r}_c)$
    \EndFor
    \State $\mat{R}_c \gets $ Low-Var-Resample($\mat{R}_c, \vec{p}(\vec{\gamma} | \mat{R}_c)$)
\EndFor
\State \textbf{return} $\mat{R}_c$
\EndProcedure
\end{algorithmic}
\end{algorithm}

In the following, we extend this approach in two ways. First, we note that the CPF formulation can include a grasped object. We achieve this by treating the grasped object as an additional link that is rigidly attached to the end-effector and sampling contact locations exclusively from the object's surface $\mathcal{S}_O$ rather than the robot surface. We refer to this method as CPFGrasp. Second, we reason over the unknown object pose $\mat{H}_O$ by introducing an additional particle filter over object pose. We call this extension Simultaneous Contact and Object Pose Estimation (SCOPE). We then implement SCOPE on a bimanual robot platform so that we can utilize the proprioceptive capability of each arm to reason over interactions, as we would see in our downstream applications. 

\subsection{Contact Particle Filter for Grasped Objects}
The algorithm for CPFGrasp is outlined in Alg.~\ref{alg:cpfgo}. To find the contact location on the surface of a grasped object, we assume that the transformation between the end-effector frame and the object frame, $\mat{H}_{O}$, is known and given as input. We also assume that an estimate of the external wrench applied to the grasped object in the end-effector frame, $\vec{\Gamma}_{E}$, is known and given as input.

Because we know $\mat{H}_{O}$, we can formulate an adjoint transformation between a given contact location $\vec{r}_{c}$ on the surface of the object $\mathcal{S}_O$ and the end-effector, $\mat{Adj}^c_E$. This transformation maps the external wrench applied in the contact frame $c$ to the equivalent wrench in the end-effector frame $E$ and is given by:

\begin{equation}
\mat{Adj}^c_{E} = 
\begin{bmatrix}
{\mat{R}^c_{E}}^T & \vec{\emptyset}_{3 \times 3} \\
-{\mat{R}^c_{E}}^T\vec{\hat{p}}^{c}_{E} & {\mat{R}^c_{E}}^T
\end{bmatrix}
\end{equation}

\noindent  We then adjust the CPF optimization presented in Eq.~\ref{eq:cpfquadoptim} accordingly: 

\begin{mini}|s|
{\vec{\alpha}}{ (\vec{\Gamma}_{E}-\mat{Adj}^{c}_{E}\vec{\Gamma}_{c})^T {\Sigma_{m}^{-1}} (\vec{\Gamma}_{E}-\mat{Adj}^{c}_{E}\vec{\Gamma}_{c})}
{}{\text{QP}(\vec{\Gamma}_{E}|\vec{r}_c)=}
\addConstraint{\vec{\alpha} \geq 0, \quad \Gamma_c = \begin{bmatrix}
\sum_{i=1}^{N_{f}} \alpha_{i} \vec{f}_{c, i} \\
\vec{\emptyset}_{3 \times N_f}
\end{bmatrix}}
\end{mini}

\noindent We solve this optimization with \textit{cvxpy} \cite{agrawal2018rewriting}.

\begin{algorithm}
\caption{CPFGrasp}\label{alg:cpfgo}
\begin{algorithmic}
\Procedure{CPFGrasp}{$\vec{\Gamma}_{E}, \mat{H}_{O}$}
\State $\mat{R}_c \gets \{\vec{r}_{c,0}, \cdots, \vec{r}_{c,N_{clp}} \} \in \mathcal{S}_{O}$ \Comment{Init particles}
\State $N_{cs} \gets $ Set Number of Iterations
\For{$i \gets 0; \; i < N_{cs}; \; i++$}
    \State $\mat{R}_c \gets $ Motion-Model$(\mat{R}_c)$
    \For{$\vec{r}_c$ in $\mat{R}_c$}
        \State $\mat{Adj}^{c}_{E} \gets \text{Adjoint}(\mat{H}_O, \vec{r}_{c})$
        \State $p(\vec{\Gamma}_{E} | \vec{r}_c) \gets \text{QP}(\vec{\Gamma}_{E} | \vec{r}_c)$
    \EndFor
    \State $\mat{R}_c \gets \text{Low-Var-Resample}(\mat{R}_c, \vec{p}(\vec{\Gamma}_{E} | \vec{r}_{c})$)
\EndFor
\State \textbf{return} $\mat{R}_c$
\EndProcedure
\end{algorithmic}
\end{algorithm}

\subsection{Simultaneous Contact and Object Pose Estimation}

SCOPE maintains a joint distribution over object poses $\mat{H}$ and contact locations $\mat{R}_c$ on the surfaces of two objects grasped by a dual-arm system as they make contact. Our approach interleaves estimating the object poses and contact locations using two complementary particle filters. 

Alg.~\ref{alg:SCOPE} shows our approach for estimating the distribution over the object poses and contact locations for a single contact event between two robots. The algorithm takes as input the estimated external wrench applied to each robot and returns a belief distribution over object poses $\mat{H}$ and contact locations $\mat{R}_c$. 

\begin{figure}
    \centering
    \includegraphics[width=0.48\textwidth]{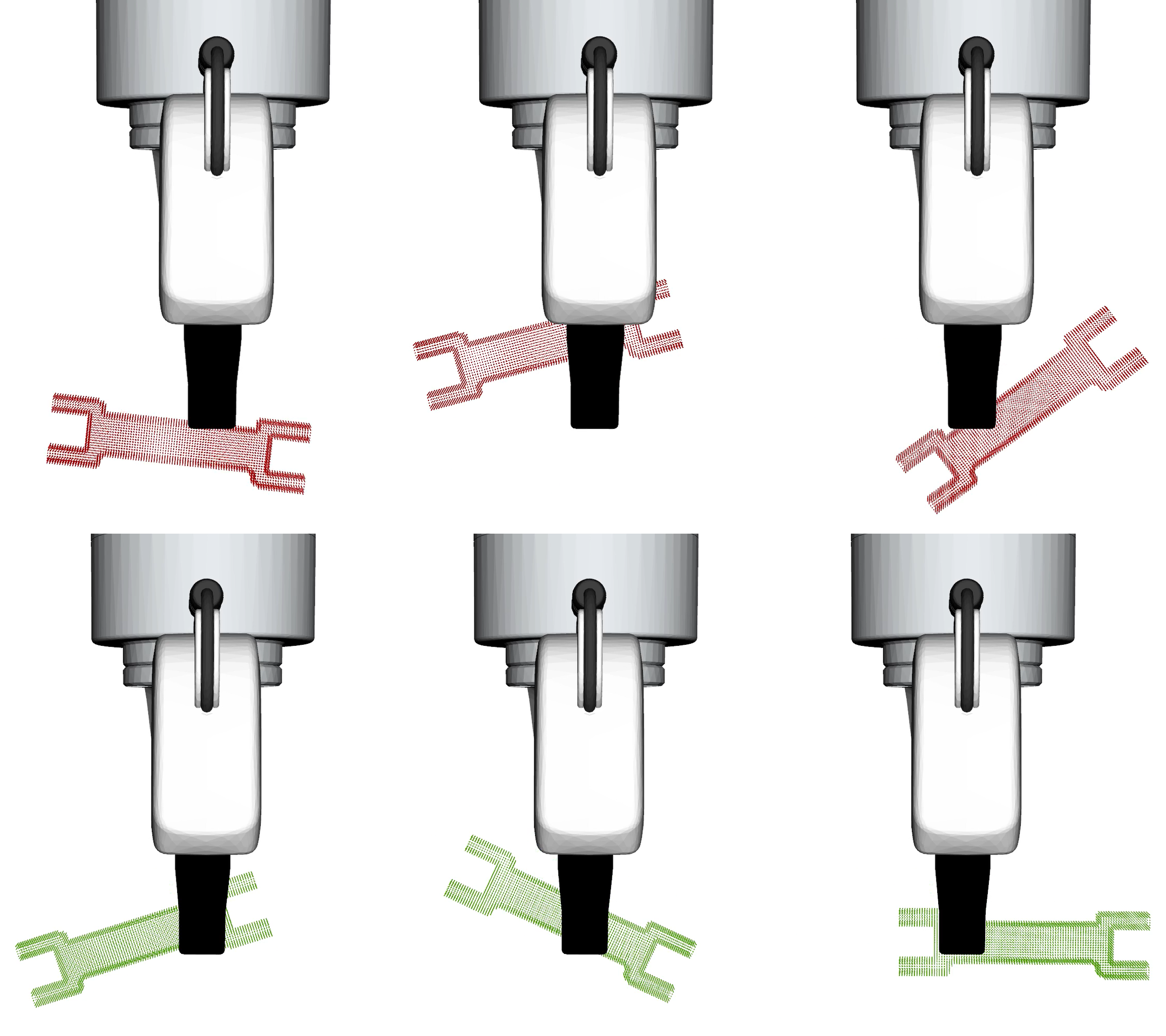}
    \caption{Examples of rejected (top) and accepted tool grasps (bottom).}
    \label{fig:grasps}
\end{figure}

\begin{algorithm}
\caption{SCOPE: $_{p} \rightarrow$ poker, $_{t} \rightarrow$ tool}
\label{alg:SCOPE}
\begin{algorithmic}
\Procedure{SCOPE}{$\vec{\Gamma}_{E,p}, \vec{\Gamma}_{E,t}$}
\State $\mat{H}_{p} \gets \{\mat{H}_{p, 0}, \cdots, \mat{H}_{p, N_{opp}} \}$
\Comment{Init pose particles}
\State $\mat{H}_{t} \gets \{\mat{H}_{t, 0}, \cdots, \mat{H}_{t, N_{opp}} \}$
\Comment{Init pose particles}
\State $\mat{H} \gets (\mat{H}_{p}, \mat{H}_{t})$
\State $N_{os} \gets $ Set Number of Iterations
\For{$i \gets 0; \; i < N_{os}; \; i++$}
    \State $\mat{H} \gets $ Noise-Model$(\mat{H})$
    \For{$j \gets 0; \; j < N_{opp}; \; j++$}
        \State $(\mat{H}_{p,j}, \mat{H}_{t,j}) \gets \mat{H}_{j}$
        \State $\mat{R}_{c,p} \gets $ CPFGrasp$(\mat{H}_{p,j}, \vec{\Gamma}_{E,p})$
        \State $\mat{R}_{c,t} \gets $ CPFGrasp$(\mat{H}_{t,j}, \vec{\Gamma}_{E,t})$
    \EndFor
    \State $\vec{S}_{OPP} \gets \text{ScoreOPPs}(\mat{H}_{p}, \mat{H}_{t}, \mat{R}_{c,p}, \mat{R}_{c,t})$
    \State $\mat{H} \gets $ Low-Var-Resample($\mat{H}, \vec{S}_{OPP}$)
\EndFor
\State \textbf{return} $\mat{H}, \mat{R}_c$
\EndProcedure
\end{algorithmic}
\end{algorithm}

Our approach randomly initializes the belief distribution over the object pose in the frame of each end-effector. As shown in Fig.~\ref{fig:grasps}, we sample transformations in $SE(2)$ that move each object in the plane of the end-effector: in our case, the plane defined by $[x, z, \theta]_E$. We accept and reject sampled transformations based on grasp validity. There are many approaches to quantifying grasp validity. In this paper, we assume that for a sampled transformation to be valid, any surface point on the transformed object must remain within 1 mm of the center of the gripper finger. For each candidate object pose in $\mat{H}$, we initialize contact locations $\mat{R}_c$ on the object surface by randomly sampling from the surface points $\mathcal{S}_{O}$.

We iteratively update the belief distribution over object poses and contact locations by scoring pairs of object particles as the sum of three losses: penetration, contact consistency, and force alignment. 

\subsubsection{Penetration Loss}

Given that our objects are rigid with known geometric models, we form a penetration loss to penalize pairs of object pose particles that are in collision:
\begin{equation}
    \label{eq:pen_loss}
    \mathcal{L}_P = max\{0, N_{PP} - \epsilon_{PP}\}
\end{equation}
\noindent where $N_{PP}$ is the number of points in penetration for a given object pose particle pair, and $\epsilon_{PP}$ is a threshold to account for many ground truth poses having a non-zero baseline level of penetration. $N_{PP}$ is calculated using the signed distance function (SDF) of the tool and the surface points of the poker. The ground truth penetration can be non-zero for two reasons: slight calibration error in our physical robot setup and coarse resolution of the voxelized object models.

\subsubsection{Contact Consistency Loss}
\label{sssec:contact_loss}
Given that the two grasped objects are in contact, we can formulate a contact consistency loss between the contact location particles of the poker $\mat{R}_{c,p}$ and of the tool $\mat{R}_{c,t}$ for each pair of object poses in $\mat{H}$:

\begin{equation}
    \label{eq:contact_loss}
    \mathcal{L}_C = \sum_{p \in \mat{R}_{c,p}, t \in \mat{R}_{c,t}}^{N_{clp}} p_{S}t_{S} \| p_{c} - t_{c} \|_2
\end{equation}

\noindent where $p$ and $t$ are the contact location particles for the poker and tool respectively. $N_{clp}$ is the number of contact location particles for each of the objects. $p_{S}$ and $t_{S}$ are the scores of each contact location particle found using Eq.~\ref{eq:cpf_score} and $p_{c}$ and $t_{c}$ are the positions of each contact location particle in the world frame.

\subsubsection{Force Alignment Loss}
Finally, we leverage the two-handedness of the interaction and the contact location belief distribution for each arm to create a metric of force alignment. We know that the forces experienced by each end-effector should be equal and opposite for this type of interaction, so we formulate an alignment loss similarly to the contact consistency loss described in Sec.~\ref{sssec:contact_loss}. Because each contact location particle in $\mat{R}_{c,p}$ and $\mat{R}_{c,t}$ has an associated external force estimate, we can write the force alignment loss as:
\begin{equation}
    \label{eq:align_loss}
    \mathcal{L}_F = \sum_{p \in \mat{R}_{c,p}, t \in \mat{R}_{c,t}}^{N_{clp}} p_{S}t_{S} \| -p_{\vec{F}} - t_{\vec{F}} \|_2
\end{equation}
where $p_{S}$ and $t_{S}$ are again the scores of each contact location particle and $p_{\vec{F}}$ and $t_{\vec{F}}$ are the external force estimates at each contact location in the world frame.

To synthesize these losses into a single score for a pair of object pose particles, we define weights $\eta_C$ and $\eta_P$ and score $S_{OPP}$:
\begin{equation}
    \label{eq:opp_score}
    S_{OPP} = \eta_P \mathcal{L}_{P} + \eta_C \mathcal{L}_{C} + \mathcal{L}_{F} 
\end{equation}
We create ${N_{opp}}^2$ object pose pairs to maximally gain information at each iteration. This is possible because each call to CPFGrasp is agnostic to the position of the other object, so we can pair each poker pose with each tool pose and vice versa. After scoring each pair using Eq.~\ref{eq:opp_score}, we compute the likelihood of each pair and resample from the best $N_{opp}$ pairs.

\section{EXPERIMENTS}

\subsection{Hardware Details}
\begin{figure}
    \centering
    \includegraphics[width=0.45\textwidth]{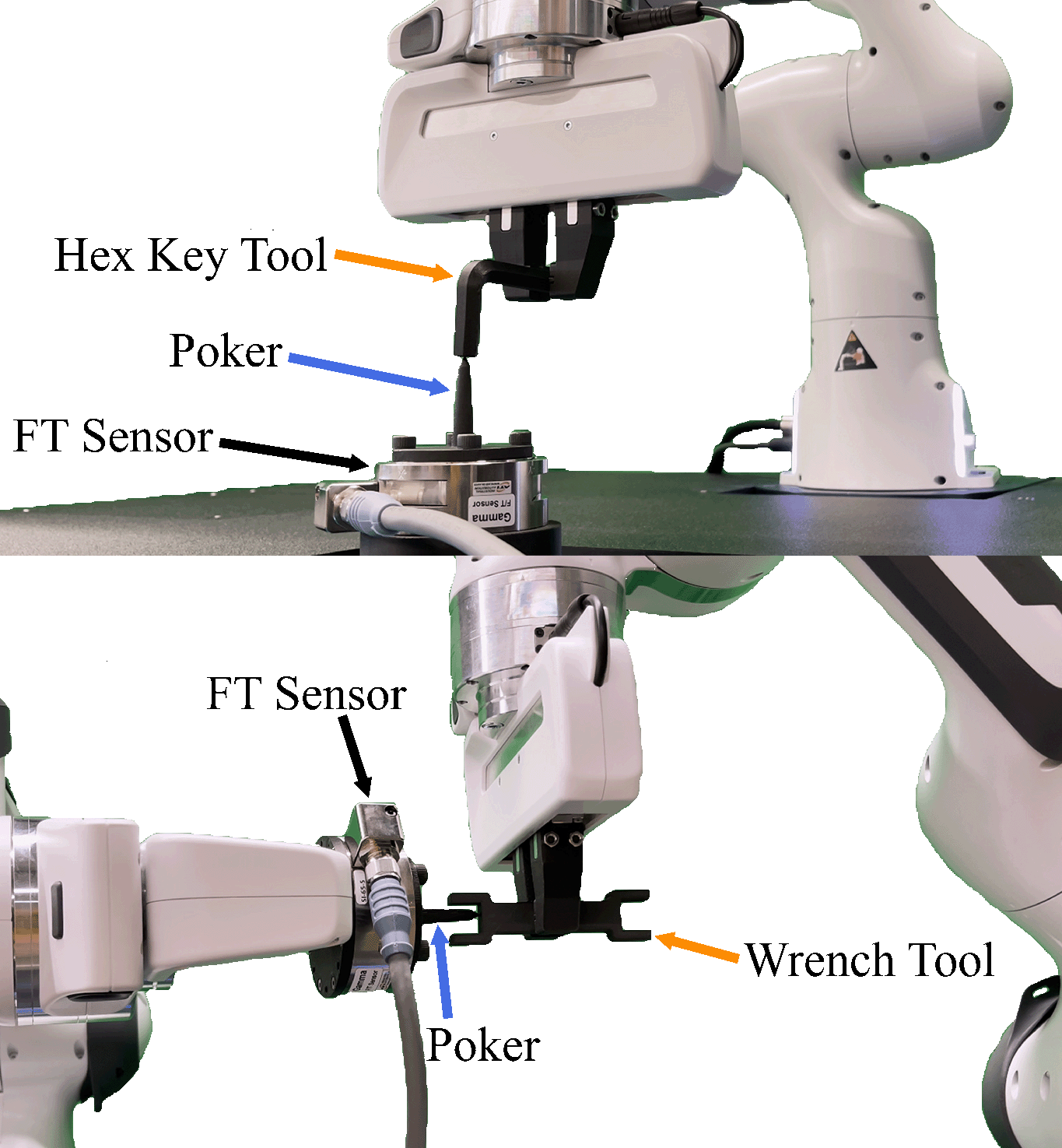}
    \caption{CPFGrasp (top) and SCOPE (bottom) experimental setups.}
    \label{fig:exp_setup}
\end{figure}

Our manipulation setup uses two Franka Emika Panda robots. To test CPFGrasp we use a single arm and to test SCOPE we use both arms. We replaced the standard two-piece knuckles from the Panda Hand with custom gripper fingers on one robot and a custom mount for an ATI Gamma FT sensor on the other, as shown in Fig.~\ref{fig:exp_setup}. We designed and 3D printed a hex key and wrench to test the CPFGrasp and SCOPE. 

The gripper fingers, hex key, and wrench are 3D printed with crosshair patterns that interface with one another. The crosshair allows us to grasp the tools in the same configuration every time, giving us a ground truth object pose to evaluate our approach. The hex key is 4 cm on the short side, 7 cm on the long side, and has an inradius of 0.5 cm. The wrench measures 10 cm in length, 2.5 cm in height, and 1 cm in width. The ground truth grasp of the hex key is 2 cm from the end of the long side while the grasp of the wrench is in line with its center of mass.

To evaluate CPFGrasp, an ATI Gamma FT sensor is rigidly attached to the table, as shown in Fig.~\ref{fig:exp_setup}. We attach a 3D printed calibration tool with the crosshair pattern to the client interface of the sensor and locate the sensor on the table. Then, we replace the calibration tool with a 3D printed poker. This allows us to measure the results of CPFGrasp against the ground truth contact location on the hex key. We use the signal from the ATI Gamma transformed into the end-effector frame for these trials. 

To evaluate SCOPE, we mount the ATI Gamma FT sensor and poker rigidly to the Panda Hand. The ATI Gamma data from the poking arm is used for these trials while the stationary arm uses its FT estimate from joint torques. It remains future work to investigate the differences in accuracy between external FT sensing and the FT estimate provided by the robot.

During data collection, the arms used impedance controllers while interacting. The arm holding the tool remained stationary while the arm with the poker moved into contact, waited, and retracted. The poking arm was stiffer than the stationary arm: it used 350 N/m in translational stiffness, 20 Nm/rad in rotational stiffness, and 70 Nm/rad in nullspace stiffness while the stationary arm used 200 N/m, 10 Nm/rad, and 0 Nm/rad respectively. All code and CAD files can be found at \url{https://github.com/MMintLab/scope}.

\subsection{Data Preparation}

\subsubsection{Object Models}
We convert the object models that we use in CPFGrasp and SCOPE from meshes to 3D voxel grids using \textit{binvox} \cite{maturana2016}. We use these voxel grids to compute 3D signed distance functions (SDFs) and their gradients using \textit{sdf-tools} \cite{armlab2021}. To obtain the point cloud of the object surface, we filter the SDF by a threshold $\epsilon_{\mathcal{S}}$. Finally, we use the gradient of the SDF to find the surface normals for each surface point. These surface normals will be used in computing the friction cone. For more efficient grasp sampling with these object models, we cap translation at $T^{x, z}_{max} =$ 3 cm and rotation at $R^{\theta}_{max} = 45^\circ$. 

On the experimental setup the poker is rigidly attached to the ATI Gamma. We map the signal from the ATI Gamma into the base frame of the robot and continue with the method defined in Alg.~\ref{alg:SCOPE}. We add a disk to the model of the poker such that the center of the disk is aligned with the center of the end-effector frame, then treat this model as a grasped object. In this way, we can reason over the pose of the poker as if it were grasped, even though it is rigidly attached.

\subsubsection{FT Signals}
One of the components of the CPF that allows it to work on physical systems is $\Sigma_{m}$, the calibrated sensor noise. We calibrate $\Sigma_{m}$ for both the ATI Gamma and the robot's FT estimate on the steady-state signal immediately before each trial. To do this, we isolate the steady-state portion of the signal then bias it to 0. We fit a Gaussian to each component in $[F_x, F_y, F_z, M_x, M_y, M_z]$. $\Sigma_{m}$ is then the diagonal matrix of the variances fitted by the Gaussians.

\section{RESULTS}

\subsection{CPFGrasp}
\begin{figure}
    \centering
    \includegraphics[width=0.48\textwidth]{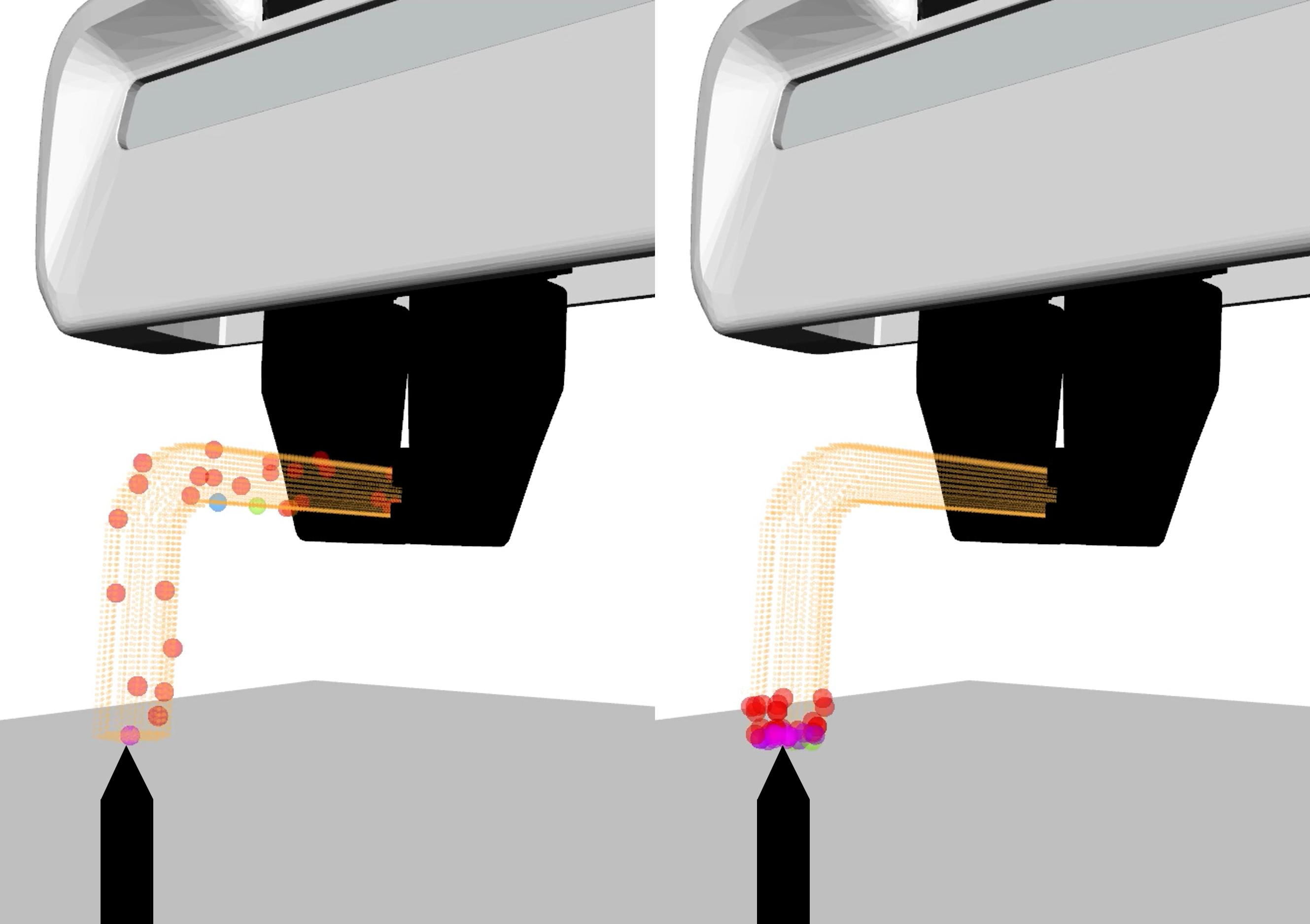}
    \caption{Initialization (left) and final distribution (right) of contact location particles for the hex key tool after 10 steps of CPFGrasp. Particles are colored with their likelihood in rainbow order: from red (low) through blue (mid) to pink (high).}
    \label{fig:cpf_hex}
\end{figure}

Over 5 trials of CPFGrasp with the hex key, the mean distance from the calibrated ground-truth contact location was 4.36 mm with a standard deviation of 0.8 mm. Fig.~\ref{fig:cpf_hex} shows how CPFGrasp is able to localize contact on the correct face of the grasped tool. Initially, the contact locations are initialized by randomly sampling from the surface of the hex key tool. After running the CPFGrasp algorithm, the contact particles converge rapidly to the contact surface between the poker and the grasped tool. This experiment validates the implementation of external contact localization given the object pose. Table~\ref{tab:cpfgrasp_params} shows the set of parameters used in this experiment.

\begin{table}
\caption{Parameters used all CPFGrasp trials.}
\label{tab:cpfgrasp_params}
\centering
\begin{tabular}{ | c | c | }
    \hline
    \multicolumn{2}{|c|}{\textbf{CPFGrasp Parameters}} \\
    \hline
    $N_{clp}$ & 40  \\
    \hline
    $N_{cs}$ & 10  \\
    \hline
    $N_f$ & 8 \\
    \hline
\end{tabular}
\end{table}

\subsection{SCOPE}

\begin{figure*}
    \centering
    \includegraphics[width=0.98\textwidth]{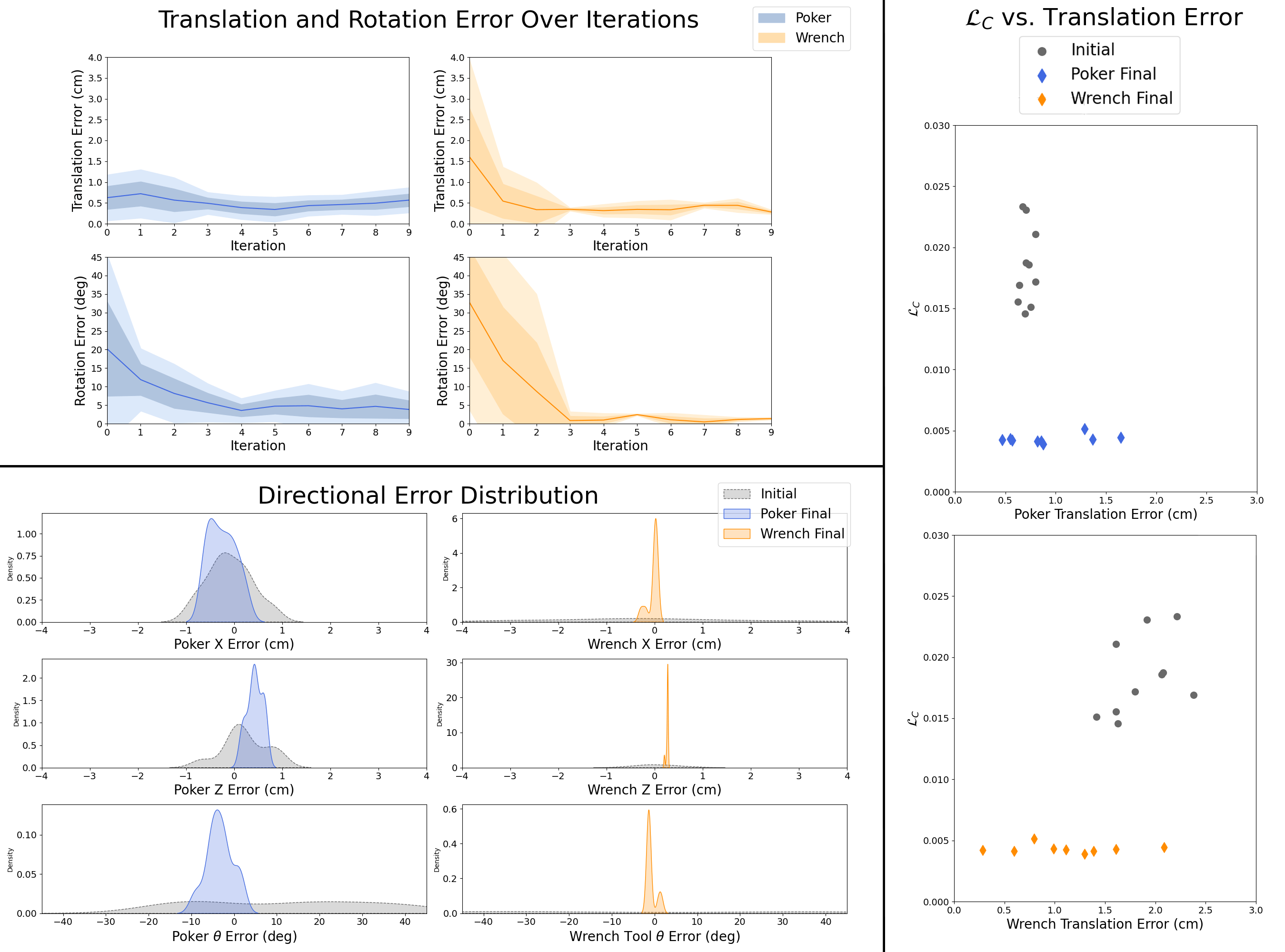}
    \caption{\textit{Top}: Mean error over 10 iterations of SCOPE for the poker and the wrench, with $\pm 2 \sigma$ shaded. \textit{Bottom}: Error in $[x, z, \theta]_E$ for the poker and the wrench with the initial distribution shaded in gray. \textit{Right:} $\mathcal{L}_{C}$ vs. translation error after using SCOPE for the poker and the wrench in 10 trials.}
    \label{fig:results}
\end{figure*}

To evaluate the efficacy of our approach, we conducted 10 trials where each trial initializes the object poses randomly. The parameters that we used for these trials are shown in Table~\ref{tab:seed_8_params}. The top pane of Fig.~\ref{fig:results} shows the convergence behavior for one of these trials. Due to the relative geometries of the objects, we note that the wrench starts with significantly higher translation error than the poker. Despite the large error and high variance of the initial distribution shown in the bottom pane, SCOPE is able to resolve object pose with low variance in just a few iterations. Referring back to the top pane, we see that the translation error of the poker and the wrench are similar after the last step. This indicates that our method has converged to a solution where the tools remain in contact with one another and are at the approximately correct poses. 

\begin{table}
\caption{Parameters used in the results displayed in Fig.~\ref{fig:opp_teaser} and Fig.~\ref{fig:results}.}
\label{tab:seed_8_params}
\centering
\begin{tabular}{ | c | c | }
    \hline
    \multicolumn{2}{|c|}{\textbf{SCOPE Parameters}} \\
    \hline
    $N_{clp}$ & 20  \\
    \hline
    $N_{opp}$ & 10 \\
    \hline
    $N_{cs}$ & 30  \\
    \hline
    $N_{os}$ & 10  \\
    \hline
    $N_f$ & 8 \\
    \hline
    $\eta_{P}$ &   0.005 \\
    \hline
    $\eta_{C}$ & 20  \\
    \hline
    $\epsilon_{PP}$ & 144 \\
    \hline
\end{tabular}
\end{table}

We note that the errors shown in Fig.~\ref{fig:results} do not converge to exactly zero. This issue is further illustrated in the right pane of Fig.~\ref{fig:results} and Table~\ref{tab:ablation}. Fig.~\ref{fig:results} shows how the contact loss is low for a distribution of translation errors. These results imply that there is an inherent ambiguity in the contact formation; i.e., when the tools are in contact and moving together along the poker axis. Our method correctly identified this ambiguity across all trials. Next, we present the effects of including our various losses in an ablation study.

\begin{table*}
\caption{Results of ablation study across loss metrics $\mathcal{L}_P, \mathcal{L}_C,$ and $\mathcal{L}_F$ over 10 trials.}
\label{tab:ablation}
\centering
\begin{tabular}{ | c | c | c | c | c | c | c | c | c | c | }
    \hline
    \multicolumn{10}{| c |}{\textbf{Ablation Study Results}} \\
    \hline
    \multirow{2}{*}{Losses} & \multicolumn{2}{c}{$\mat{H}_{p}$ Trans Error (cm) $\downarrow$} \vline & \multicolumn{2}{c}{$\mat{H}_{p}$ Rot Error (deg) $\downarrow$} \vline & \multicolumn{2}{c}{$\mat{H}_{t}$ Trans Error (cm) $\downarrow$} \vline & \multicolumn{2}{c}{$\mat{H}_{t}$ Rot Error (deg) $\downarrow$} \vline & \multirow{2}{*}{$E_{agg} \downarrow$} \\
    \cline{2-9}
    & Mean & Stdev & Mean & Stdev & Mean & Stdev & Mean & Stdev & \\
    \hline
    $\mathcal{L}_P$  & 0.71 & 0.13 & 26.68 & 5.12 & 1.88 & 0.93 & 25.57 & 11.59 & 7.27 \\
    \hline
    $\mathcal{L}_C$ & 0.70 & 0.17 & 12.03 & 1.79 & 0.80 & 0.12 & 14.32 & 1.52 & 3.85 \\
    \hline
    $\mathcal{L}_F$ & 0.88 & 0.19 & 21.13 & 1.90 & 1.78 & 0.15 & 6.64 & 1.25 & 5.18 \\
    \hline
    $\mathcal{L}_P, \mathcal{L}_C$ & 0.93 & 0.21 & 12.83 & 2.61 & 0.94 & 0.21 & 11.11 & 1.55 & 4.02 \\
    \hline
    $\mathcal{L}_C, \mathcal{L}_F$ & 0.95 & 0.15 & 11.76 & 1.64 & 0.85 & 0.07 & 7.82 & 1.05 & 3.56 \\
    \hline
    $\mathcal{L}_P, \mathcal{L}_F$ & 0.91 & 0.21 & 22.64 & 1.95 & 1.79 & 0.13 & 7.37 & 1.15 & 5.42 \\
    \hline
    $\mathcal{L}_P, \mathcal{L}_C, \mathcal{L}_F$ & 1.03 & 0.15 & 4.94 & 1.76 & 1.16 & 0.13 & 6.44 & 1.48 & 3.20 \\
    \hline
\end{tabular}
\end{table*}

\begin{table*}
\caption{Results of parameter sweep across $N_{clp}$ and $N_{opp}$  over 10 trials.}
\label{tab:paramsweep}
\centering
\begin{tabular}{ | c | c | c | c | c | c | c | c | c | c | c | }
    \hline
    \multicolumn{11}{| c |}{\textbf{Hyperparameter Sweep Results}} \\
    \hline
    \multirow{2}{*}{$N_{clp}$} & \multirow{2}{*}{$N_{opp}$} & \multicolumn{2}{c}{$\mat{H}_{p}$ Trans Error (cm) $\downarrow$} \vline & \multicolumn{2}{c}{$\mat{H}_{p}$ Rot Error (deg) $\downarrow$} \vline & \multicolumn{2}{c}{$\mat{H}_{t}$ Trans Error (cm) $\downarrow$} \vline & \multicolumn{2}{c}{$\mat{H}_{t}$ Rot Error (deg) $\downarrow$} \vline & \multirow{2}{*}{$E_{agg} \downarrow$} \\
    \cline{3-10}
    & & Mean & Stdev & Mean & Stdev & Mean & Stdev & Mean & Stdev & \\
    \hline
    10 & 5 & 0.92 & 0.14 & 15.08 & 1.59 & 1.61 & 0.03 & 15.57 & 0.40 & 5.27 \\
    \hline
    10 & 10 & 0.90 & 0.17 & 8.74 & 1.76 & 1.03 & 0.10 & 18.75 & 1.22 & 4.37 \\
    \hline
    20 & 5 & 0.92 & 0.13 & 11.08 & 1.47 & 1.00 & 0.082 & 12.22 & 0.35 & 4.00 \\
    \hline
    20 & 10 & 1.03 & 0.15 & 4.94 & 1.76 & 1.16 & 0.13 & 6.44 & 1.48 & 3.20 \\
    \hline
\end{tabular}
\end{table*}

\subsection{Ablation}

To validate our choice of loss metrics, we conducted an ablation study across $\mathcal{L}_{P}, \mathcal{L}_{C},$ and $\mathcal{L}_{F}$. We conducted this study with 10 trials for each combination of losses. Results for this study are shown in Table~\ref{tab:ablation}. For an interpretable error metric, we define aggregate error $E_{agg}$ in terms of translation error $E_{\mat{H}_O, t}$, rotational error $E_{\mat{H}_O, r}$, and radius of gyration $r_{G,O}$. $E_{agg}$ measures the average error across the translation and orientation of both objects:
\begin{equation}
    \label{eq:agg_error}
    E_{agg} = E_{\mat{H}_{p}, t} + E_{\mat{H}_{t}, t} + r_{G, p}E_{\mat{H}_{p}, r} + r_{G, t}E_{\mat{H}_{t}, r}
\end{equation}
The radii of gyration are 5.25 cm and 5 cm for the poker and wrench respectively. 

As we examine Table~\ref{tab:ablation}, we notice several interesting themes. As a standalone loss metric, contact consistency loss outperforms both penetration loss and force alignment loss and is comparable to the combination of all three loss metrics. Adding force alignment loss to the contact consistency loss is also comparable to the combination of all three loss metrics. 

There are several loss combinations that provide more accurate estimates in one category but much worse estimates in another category, such as contact consistency. This loss metric has the lowest translation error for both the poker and the wrench across all combinations, but has high rotational error for both of the parts. Without implementing the force alignment loss, there is no information that SCOPE can use about how the objects are interacting with one another to resolve the rotational error.

\subsection{Hyperparameter Sweep}

To explore how changing the parameters impacts our method, we conducted a hyperparameter sweep across the $N_{clp}$ and $N_{opp}$. For each set of parameters, we conducted 10 trials. Results are shown in Table~\ref{tab:paramsweep}. To provide a clear success metric, we again use $E_{agg}$ as defined in Eq.~\ref{eq:agg_error}. These results show that larger $N_{clp}$ and $N_{opp}$ lead to better SCOPE results. 

\section{FUTURE WORK}
Building on the success of CPFGrasp and SCOPE, we look ahead to multi-action framework, next-action selection, and multi-modality integration as relevant extensions of our method to increase performance as well as applicability to downstream tasks.

\textbf{Multi-Action Framework:} The persistence of ambiguity in object pose after a single interaction begs the question of how best to alleviate this uncertainty over multiple interactions. We hypothesize that adding a memory loss term to our scoring function (Eq.~\ref{eq:opp_score}) in order to reason over multiple contact events will allow us to further reduce the uncertainty in our object pose estimation. 

\textbf{Next-Action Selection:} With a multi-action framework of our method, we gain the ability to choose the next action that we take to disambiguate object pose. We can leverage the object pose distributions that SCOPE provides in order to choose actions that act along the dimensions with most uncertainty.

\textbf{Multi-Modality Integration:} As mentioned in Sec.~\ref{sec:intro} and Sec.~\ref{sec:related_work}, our method is complementary to other types of manipulation approaches. In particular, our method is robust to the occlusions that limit visual approaches. However, in non-occluded scenes, visual approaches could be utilized in tandem with SCOPE to increase performance. For example, we could use visual feedback to initialize SCOPE rather than the random initialization we currently use. We could also use visual approaches to identify grasped objects in order to relax our assumption of known object models. 

\section{CONCLUSION}

In this paper, we demonstrated the efficacy of CPFGrasp in estimating contact location on a known object with known pose. Further, we evaluated SCOPE in simultaneously estimating both contact location and object pose for each object grasped by a dual-arm robot. Both CPFGrasp and SCOPE use exclusively proprioception and force-torque signals which sets them apart from most current robotic manipulation approaches. We find that SCOPE solves for accurate object poses and contact locations that maintain the context of an interaction, even for small and non-convex tool geometries. 

\section{ACKNOWLEDGEMENT}
The authors thank Peter Mitrano and Mark van der Merwe for their contributions to the robot control interface.


\bibliographystyle{IEEEtran}
\bibliography{IEEEabrv,ref}

\end{document}